\documentclass[10pt,twocolumn,letterpaper]{article}

\PassOptionsToPackage{numbers, square}{natbib}

\usepackage{cvpr}              

\usepackage[utf8]{inputenc} 
\usepackage[T1]{fontenc}    
\usepackage{booktabs}       
\usepackage{amsfonts}       
\usepackage{nicefrac}       
\usepackage{microtype}      
\usepackage{amsmath}
\usepackage{pifont}
\usepackage{caption}
\usepackage{graphicx}
\usepackage{multirow}
\definecolor{cvprblue}{rgb}{0.21,0.49,0.74}
\usepackage[pagebackref,breaklinks,colorlinks,allcolors=cvprblue]{hyperref}
\usepackage[accsupp]{axessibility}
\def\paperID{*****} 
\def\confName{CVPR}
\def\confYear{2026}
\DeclareMathOperator{\arctantwo}{arctan2}
\newcommand{\commentout}[1]{}
\newcommand*\samethanks[1][\value{footnote}]{\footnotemark[#1]}

\title{RQR3D: Reparametrizing the regression targets for BEV-based 3D object detection}

\author{%
  Ozsel Kilinc \thanks{Equal contribution.} \thanks{This work was done prior to joining Amazon.} \\
  Amazon Lab 126 \\
  Cambridge, UK \\
  {\tt\small ozsel@amazon.co.uk} \\
  \and
  Cem Tarhan\samethanks[1] \\
  Togg/Trutek AI Team\\
  Ankara, Turkey \\
  {\tt\small cem.tarhan@togg.com.tr} \\
}

\begin{document}

\maketitle

\begin{abstract}

Accurate, fast, and reliable 3D perception is essential for autonomous driving. Recently, bird's-eye view (BEV)-based perception approaches have emerged as superior alternatives to perspective-based solutions, offering enhanced spatial understanding and more natural outputs for planning. Existing BEV-based 3D object detection methods, typically using an angle-based representation, directly estimate the size and orientation of rotated bounding boxes. We observe that BEV-based 3D object detection is analogous to aerial oriented object detection, where angle-based methods are known to suffer from discontinuities in their loss functions. Drawing inspiration from this domain, we propose \textbf{R}estricted \textbf{Q}uadrilateral \textbf{R}epresentation to define \textbf{3D} regression targets. RQR3D regresses the smallest horizontal bounding box encapsulating the oriented box, along with the offsets between the corners of these two boxes, thereby transforming the oriented object detection problem into a keypoint regression task. We employ RQR3D within an anchor-free single-stage object detection method achieving state-of-the-art performance. We show that the proposed architecture is compatible with different object detection approaches. Furthermore, we introduce a simplified radar fusion backbone that applies standard 2D convolutions to radar features. This backbone leverages the inherent 2D structure of the data for efficient and geometrically consistent processing without over-parameterization, thereby eliminating the need for voxel grouping and sparse convolutions. Extensive evaluations on the nuScenes dataset show that RQR3D achieves SotA camera-radar 3D object detection performance despite its lightweight design, reaching 67.5 NDS and 59.7 mAP with reduced translation and orientation errors, which are crucial for safe autonomous driving. 

\end{abstract}

\section{Introduction}

Autonomous driving necessitates comprehensive environmental perception, encompassing dynamic elements such as vehicles, bicycles, and pedestrians, as well as an understanding of static conditions like road topology. Effective perception systems must integrate these diverse aspects to ensure safe and efficient navigation. In recent years, bird's-eye-view (BEV)-based approaches have become increasingly prominent in both academic and industrial settings. This shift is driven by BEV's superior spatial understanding and more intuitive outputs, which are crucial for advanced perception systems in autonomous driving. The 2D to BEV projection constitutes a pivotal component within these systems, facilitating the seamless integration of rich visual data with spatial context. This synthesis significantly enhances the precision of object detection, tracking, and scene comprehension, thereby augmenting the overall efficacy and dependability of autonomous driving technologies. 

There exist two principal paradigms for 2D to BEV projection. The foundational works, SimpleBEV \cite{SimpleBEV} and M2BEV \cite{M2BEV}, utilize Inverse Perspective Mapping (IPM), which employs geometric back-projection to map each BEV voxel to its corresponding coordinates in the image plane. More advanced IPM variants leverage multi-layer perceptrons (MLPs) to refine the feature transformation \cite{MLP}. Conversely, Lift-Splat (LS)-based approaches implement a 'push'-based strategy, wherein each image pixel is associated with a probabilistic depth distribution, enabling it to contribute to multiple voxels in 3D space. The seminal Lift-Splat-Shoot (LSS) framework \cite{LSS} employs a depth estimation network. A more sophisticated implementation is presented in BEVDepth \cite{li2023bevdepth}, which incorporates high-dimensional camera calibration embeddings via linear layers. Transformer-based approaches diverge from these two paradigms, projecting image features to BEV representation through the utilization of learnable BEV queries and spatial cross-attention layers \cite{li2024bevformer, wang2022detr3d}.

BEV representation also leverages the implementation simplicity of early sensor fusion. Point cloud (PC) sensors, such as radar and LiDAR, can be seamlessly integrated with projected image features. Recent advancements in camera-radar fusion \cite{CRN, CRT, Hydra, RCB} have addressed the inherent sparsity of radar by adopting sophisticated fusion architectures, including temporal modules, deformable attention mechanisms, and LiDAR-inspired radar encoders. These developments have positioned this sensor combination as a cost-effective and scalable alternative to LiDAR for 3D object detection. The synergy between camera and radar technologies facilitates robust environmental perception, offering a viable solution that balances performance with affordability.

After BEV representation is processed, and optionally fused, it is subsequently shared among task-specific heads, such as 3D object detection, segmentation, prediction, occupancy, and road topology. There exists a substantial body of research in BEV-based 3D object detection literature, including \cite{li2024bevformer, wang2022detr3d, yang2023bevformer, huang2021bevdet, huang2022bevdet4d, li2023bevdepth, liu2022petr, liu2023petrv2, HOP, RayDN, SparseBEV}. These methods typically adopt angle-based target regressions proposed by CenterPoint \cite{yin2021center} for the 3D detection head. Although angle-based methods are noted in the 2D oriented object detection literature for suffering from loss function discontinuities \cite{yang2020arbitrary}, CenterPoint addresses this issue by introducing the sine and cosine of the yaw angle as regression targets. However, learning the size of the bounding box, which is inherently rotation-invariant in angle-based representation, continues to pose challenges due to the rotation-variant nature of convolutions. CenterPoint mitigates this issue by introducing class-specific regression heads, which are refined using dataset-specific heuristics.

We observe that BEV-based 3D object detection is analogous to the well-studied domain of aerial oriented object detection with additional z-axis targets. Drawing inspiration from this field, we propose the \textbf{R}estricted \textbf{Q}uadrilateral \textbf{R}epresentation to define \textbf{3D} regression targets. RQR3D regresses the smallest horizontal bounding box encapsulating the oriented box, along with the offsets between the corners of these two boxes. By using only two out of four offsets, we restrict the representation to allow only rectangular cuboids with right angles. This proposed representation transforms the oriented object detection problem into a keypoint regression task, which is more amenable to learning using translation-invariant convolutions. RQR3D can be integrated with any 3D object detection approach. We modify an anchor-free single-stage 2D object detection method for BEV-based 3D object detection, utilizing RQR3D regression targets. Additionally, we integrate an objectness head into this detector to address the class imbalance issue. Furthermore, we present a lightweight radar fusion backbone that eliminates the need for voxel grouping and processes the BEV-mapped PC using conventional 2D convolutions instead of sparse convolutions. Extensive evaluations on the nuScenes dataset reveal that RQR3D positions itself among SotA methods in camera-radar 3D object detection scoring  67.5 in NDS and 59.7 in mAP, with notable advancements in translation and orientation accuracy. These consistent improvements highlight the robustness, accuracy, and practical applicability of our approach.


\section{Related Work}

\paragraph{Aerial Object Detection.}

As opposed to typical object detectors that utilize horizontal bounding boxes, targets in aerial images are often arbitrarily oriented and densely packed. Approaches in this domain generally evolve from classical object detectors by incorporating orientation regression \cite{pan2020dynamic, xie2021oriented, han2021redet}. Angle-based methods, which define an oriented bounding box as \((x_{ctr}, y_{ctr}, w, h, \theta)\), (center, size, and orientation angle), are known to suffer from boundary discontinuity issues \cite{zhao2024projecting}. Yang et al. \cite{yang2020arbitrary} address this by transforming angular regression into a classification task. Quadrilateral methods define an oriented bounding box using an outer horizontal box and offsets from its four corners, denoted as \((x_{ctr}, y_{ctr}, w, h, u1, v1, u2, v2)\). While these methods offer greater flexibility in representing objects' shapes and orientations, they also face challenges with ambiguous definitions. \cite{li2022oriented} propose an effective representation using adaptive points, and \cite{zhao2024projecting} combine point set and axis representations.

\paragraph{BEV-based 3D Object Detection.}

BEV-based 3D object detection resembles 2D oriented object detection but incorporates additional z-axis targets.  In this context, orientation is considered along the z-axis, defining a 3D oriented bounding box as \((x_{ctr}, y_{ctr}, z_{ctr}, w, l, h, \theta)\), where \(\theta\) represents the yaw angle. Although anchor-based methods have been explored, CenterPoint \cite{yin2021center} has demonstrated that center-based representations excel in learning rotational invariance and object equivalence. This approach has surpassed anchor-based methods in performance, establishing itself as the preferred output representation for BEV-based 3D object detection. CenterPoint uses an angle-based representation defined as \((x_{svlr}, y_{svlr}, z_{bottom}, w, l, h, \sin(\theta), \cos(\theta))\), where \((x_{svlr}, y_{svlr})\) denotes sub-voxel location refinement, \(z_{bottom}\) indicates height above ground, and orientation is parameterized by its sine and cosine to address boundary discontinuity issues. The 3D size \(w, l, h\) in this representation are inherently rotation-invariant. However, learning these parameters remains challenging due to the rotation-variant nature of convolutions. CenterPoint mitigates this issue by employing class-specific regression heads, each refined using dataset-specific heuristics, such as the expected size of a class. This approach, while effective, constrains the generalization capacity.

\paragraph{Radar Processing Methods.} PC processing is crucial for 3D perception with LiDAR and radar sensors. LiDAR provides rich 3D data, while radar mainly offers depth and lateral data, lacking reliable height measurements. SECOND \cite{SECOND} introduced a 3D sparse convolutional neural network for LiDAR PC processing, using voxel-based representation to extract geometric features and enable spatial encoding. PointPillars \cite{PointPillars} collapses the height axis to project PC data into a pseudo-image format for faster 2D convolutional processing, sacrificing spatial fidelity and generally underperforming compared to SECOND in detection accuracy. PillarNet \cite{PillarNet} aggregates points into 2D BEV-aligned pillar structures and applies sparse convolutions, balancing efficiency with enhanced feature extraction. These methods require: (1) point-to-voxel collection to discretize the sparse input space, and (2) sparse convolutional operations for efficient processing. These operations depend on hardware-specific optimizations in specialized libraries for deployment.

\paragraph{Camera-only and Camera-Radar-based 3D Object Detection Methods.}
SOLOFusion \cite{SOLO} redefines camera-only 3D object detection as a multi-view stereo matching task, combining long-term, low-resolution fusion with short-term, high-resolution stereo for better localization. StreamPETR \cite{StreamPETR} proposes a temporal modeling approach, propagating object queries across frames with a memory queue. It avoids dense BEV reducing temporal attention costs. HoP \cite{HOP} improves temporal learning by predicting historical object states with decoders. SparseBEV \cite{SparseBEV} uses learnable query pillars and adaptive attention for camera-only 3D detection. RayDN \cite{RayDN} improves DETR-style detectors by reducing false positives with depth-aware hard negative sampling along camera rays during training. CRT-Fusion \cite{CRT} is a camera-radar fusion framework for 3D object detection that uses predicted object motion and occupancy to align BEV features across frames. CRN \cite{CRN} implements detection, tracking, and segmentation by transforming perspective-view image features into BEV using radar guidance and multi-modal deformable attention. HyDRa \cite{Hydra} combines radar and image features in both perspective and BEV spaces for depth-aware 3D perception that enhances depth estimation and spatial alignment. RCBEVDet \cite{RCB} performs camera-radar fusion in BEV space, using radar cross-section (RCS) information for better spatial representation. It aligns camera and radar features via deformable attention. RICCARDO \cite{riccardo} leverages motion-compensated multi-sweep radar accumulation for temporal stability rather than recurrent fusion. Its multi-stage inference makes it complex and compute-heavy, despite not using transformers. RaCFormer \cite{racformer} leverages temporal fusion through multi-frame radar and camera sequences, with a ConvGRU to propagate motion cues, making it one of the most computationally demanding DETR3D-style camera–radar fusion frameworks to date.

\begin{figure*}[t]
\centerline{\includegraphics[width=0.90\linewidth]{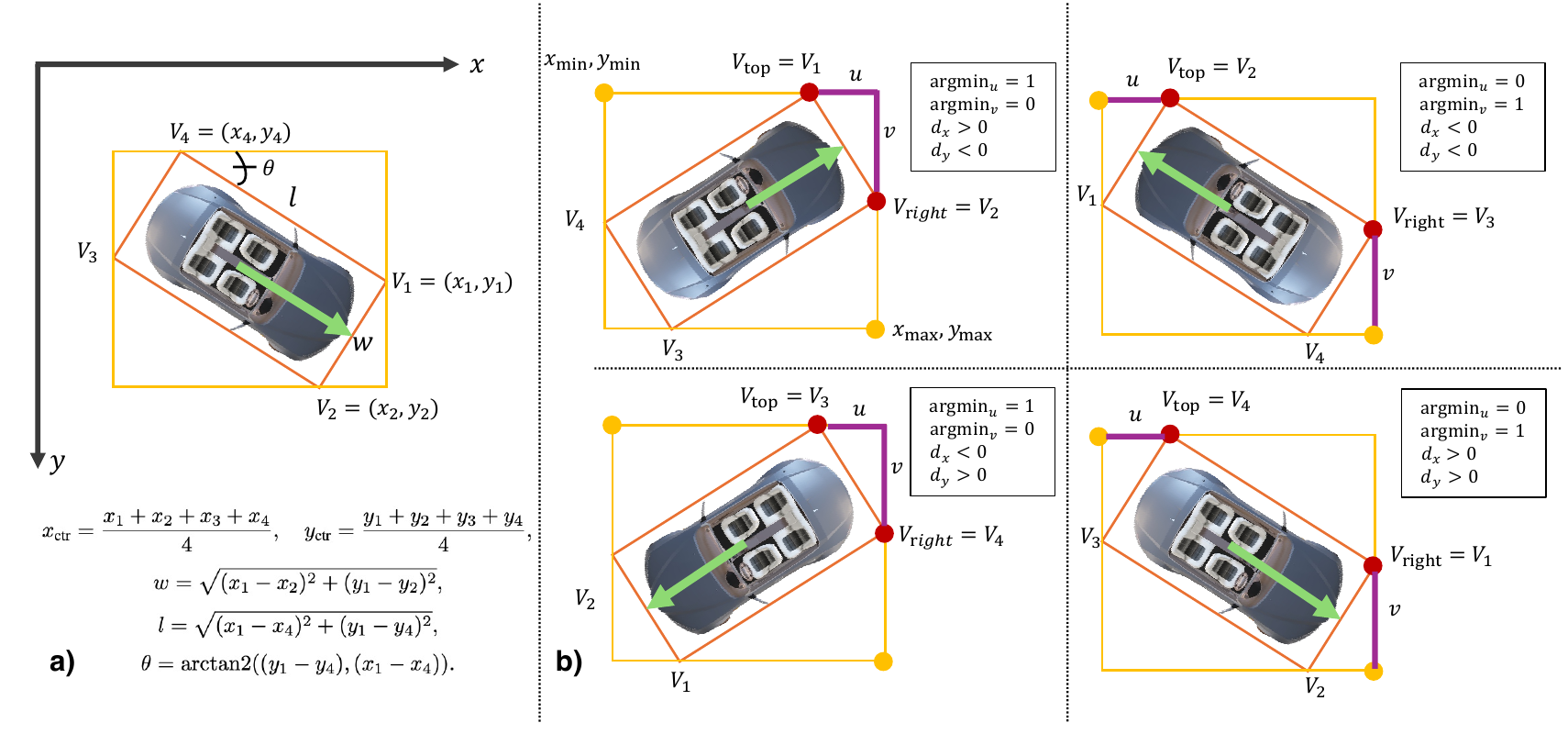}}
\caption{a) First, we identify the smallest axis-aligned 2D bounding box (yellow) that encapsulates the oriented box (orange). We then determine the offsets from the topmost and rightmost corners of the orange box to the corresponding corners of the yellow box. 
b) The shortest offsets on x-axis and y-axis are assigned as \((u,v)\), with their indices denoted as \((\arg\min_u, \arg\min_v)\), respectively. 
A representation with these parameters limits the orientation within a \(180^\text{o}\) angular range. There are four variations depending on the quadrant of the orientation, but two of these cases cannot be differentiated. To address this problem we use the distance from the center of the \(V_1V_2\) edge to the bounding box center, \((d_x, d_y)\), to estimate the orientation.
}
\label{fig:reparam}
\end{figure*}

\section{Methodology}

\subsection{Problem Definition}

Given a 3D object bounding box defined in the bird's eye view as a rectangular cuboid in which all of its dihedral angles are right angles. It can be rotated along the z-axis, i.e. the yaw angle; but not along the x- and y-axes. Let \((V_1, \dots, V_8)\) be its vertices whose 3D coordinates are denoted as \((x_1, y_1, z_\text{bottom}), \dots, 
(x_4, y_4, z_\text{top})\), respectively. Existing BEV-based 3D object detection methods generally employ angle-based approach and aim to regress a 3D center \((x_{\text{ctr}}, y_{\text{ctr}}, z_{\text{ctr}})\), a 3D size \((w, l, h)\), and a rotation expressed by the yaw angle \(\theta\). 

Noting that, aside from the regressions of \(z_{\text{ctr}}\) and \(h\), the remainder of the problem closely resembles the 2D oriented object detection addressed in the aerial imagery literature, where angle-based approaches are known to suffer from loss function discontinuities leading to learning instabilities. We draw inspiration from this literature and propose a restricted quadrilateral representation for the regression targets related to the x- and y-axes, as detailed below. 

\subsection{Restricted Quadrilateral Representation for 3D Bounding Boxes}
\label{rqr3d}
Let \(\mathcal{B}\) represent the bottom face of the given 3D bounding box, with \((V_1, V_2, V_3, V_4)\) as its vertices. We disregard the z-axis of the vertices since they all lie on the \(z=z_{bottom}\) plane. First, we find the smallest axis-aligned 2D bounding box \(\mathcal{C}\) that encapsulates \(\mathcal{B}\). As \(\mathcal{C}\) is horizontal, it can be defined by its top-left and bottom-right corners, \((x_{\text{min}}, y_{\text{min}})\) and \((x_{\text{max}}, y_{\text{max}})\), which can be determined as follows: 

\begin{equation}
\begin{aligned}
(x_{\min},\, y_{\min}) &= (\min_{1\le i\le 4} x_i,\, \min_{1\le i\le 4} y_i),\\
(x_{\max},\, y_{\max}) &= (\max_{1\le i\le 4} x_i,\, \max_{1\le i\le 4} y_i).
\end{aligned}
\end{equation}

Then, we use the offsets from the corners of \(\mathcal{B}\) to the corners of \(\mathcal{C}\). Since the corners are right angles and symmetric with respect to (\(x_\text{ctr}, y_\text{ctr}\)), it suffices to restrict the general quadrilateral representation and find the distances for two adjacent corners \(\mathcal{B}\). Thus, we define \(V_\text{top}=V_{\arg\min_y}\) and \(V_\text{right}=V_{\arg\max_x}\), and correspondingly \(x_\text{top}=x_{\arg\min_y}\) and \(y_\text{right}=y_{\arg\max_x}\) where 

\begin{equation}
\begin{pmatrix}
\arg\max_x \\
\arg\min_y
\end{pmatrix}
= 
\begin{pmatrix}
\arg\max_{1\le i\le 4} x_i\\[2pt]
\arg\min_{1\le i\le 4} y_i
\end{pmatrix}.
\end{equation}

Depending on the orientation of the bounding box, \(V_\text{top}\) and \(V_\text{right}\) can be any two adjacent corners of \(\mathcal{B}\). We calculate the distance from \(x_\text{top}\) to both \(x_\text{min}\) and \(x_\text{max}\), as well as the distance from \(y_\text{right}\) to both \(y_\text{min}\) and \(y_\text{max}\). We define the minimum of each of these distances as \(u\) and \(v\), and denote their indices as \(\arg\min_u\) and \(\arg\min_v\) as follows:

\begin{equation}
    \begin{pmatrix}
    u \\
    v \\
    \arg\min_u \\
    \arg\min_v
    \end{pmatrix}
    =
    \begin{pmatrix}
    \min (x_{\text{top}} - x_{\text{min}}, & x_{\text{max}} - x_{\text{top}}) \\
    \min (y_{\text{right}} - y_{\text{min}}, & y_{\text{max}} - y_{\text{right}}) \\
    \arg\min(x_{\text{top}} - x_{\text{min}}, & x_{\text{max}} - x_{\text{top}}) \\
    \arg\min(y_{\text{right}} - y_{\text{min}}, & y_{\text{max}} - y_{\text{right}})
    \end{pmatrix}
    \renewcommand{\arraystretch}{1.0} 
\end{equation}

One might propose using \((x_{\text{top}} - x_{\text{min}})\) and \((y_{\text{max}} - y_{\text{right}})\) directly; however, this choice results in dramatic jumps in the regression values and loss discontinuities when the orientation is close to the cardinal angles. Therefore, we use the \(\min\) operator to create a symmetric regression target around the cardinal angles, transferring these jumps to \(\arg\min_u\) and \(\arg\min_v\), which are either 0 or 1. We observe that this approach helps stabilize the training and yields better results.  

Now, a representation with these parameters limits \((x_\text{ctr}, y_\text{ctr}, w, l,\theta)\)  within a \(180^\text{o}\) angular range. To address this problem we introduce two additional regression targets. We define the distance from the center of the \(V_1V_2\) edge to the bounding box center as 

\begin{equation}
    d_x = \frac{x_1 + x_2}{2} - x_{\text{ctr}}, \quad
    d_y = \frac{y_1 + y_2}{2} - y_{\text{ctr}}, 
\end{equation}
and determine the orientation using \((d_x, d_y)\). We maintain \(z_{\text{ctr}} = \nicefrac{(z_\text{bottom} + z_\text{top})}{2}\) and \(h = {z_\text{top} - z_\text{bottom}}\). Finally, RQR3D uses \((x_{\text{min}}, y_{\text{min}}, x_{\text{max}}, y_{\text{max}})\) as the bounding box regression target and \((u, v, \arg\min_u, \arg\min_v, d_x, d_y, z_{\text{ctr}}, h)\) as the keypoint targets. Although the keypoint targets exhibit slight redundancy, empirical observations have not indicated any adverse effects. We provide comprehensive details on obtaining \((x_\text{ctr}, y_\text{ctr}, w, l,\theta)\) and \((V_1, \dots, V_8)\) from the RQR3D representation in the Appendices. 

\begin{figure*}[t]
\centerline{\includegraphics[width=0.90\linewidth,trim={0.9cm 5.3cm 8cm 4.4cm},clip]{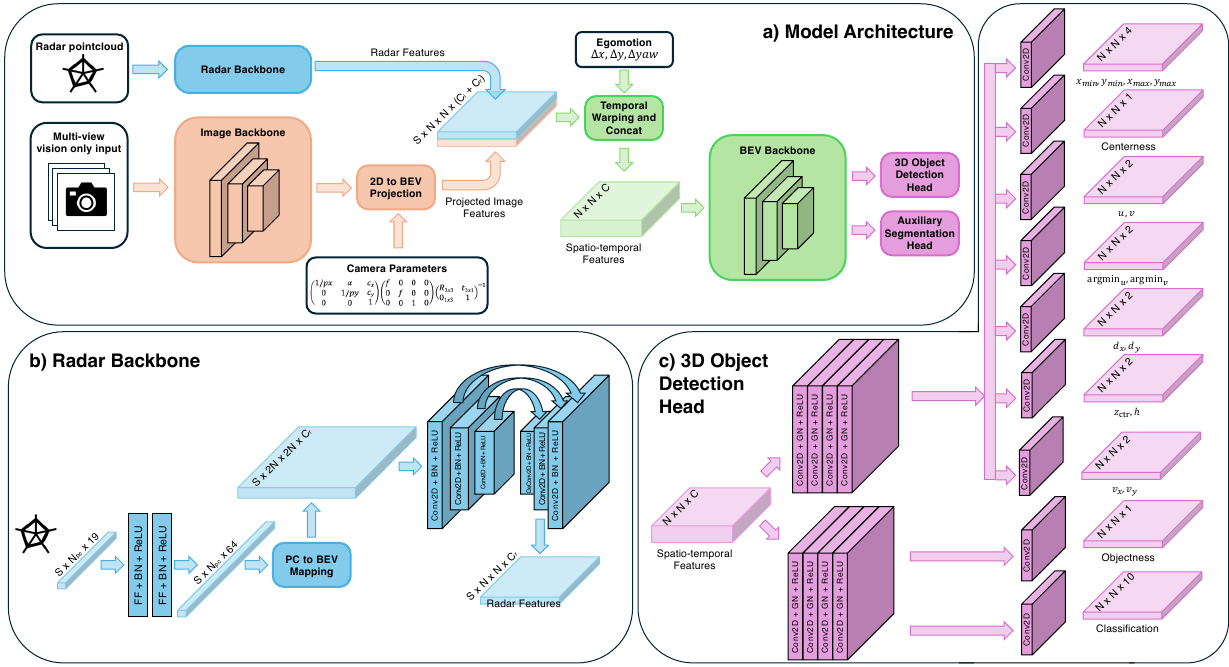}}
\caption{
a) The overall model architecture, comprising an image backbone, radar backbone, 2D to BEV projection module, temporal module, BEV backbone, and task-specific heads.
b) The simplified radar processing backbone utilizing feedforward layers, PC to BEV mapping, and standard 2D convolutions, as detailed in Section \ref{radar_bb}. 
c) The 3D object detection head with objectness, as introduced in Section \ref{3dhead}, employing the regression targets proposed in Section \ref{rqr3d}.   
}
\label{fig:block_diag}
\end{figure*}

\subsection{3D Object Detection Head}
\label{3dhead}
We utilize the anchor-free, single-stage FCOS model \cite{tian2019fcos}, adapting it for enhanced performance in BEV-based object detection. FCOS predicts a 4D vector that encodes the relative distances to each edge (\textit{top, left, right, bottom}) at every foreground pixel, derived from \((x_{\text{min}}, y_{\text{min}}, x_{\text{max}}, y_{\text{max}})\) and the pixel's location. Additionally, FCOS estimates centerness for each foreground pixel, representing its normalized distance from the object center, and integrates this with the classification score. For optimization, we employ focal loss \cite{lin2017focal} for classification, binary cross-entropy loss for centerness as in \cite{tian2019fcos}, and generalized IOU loss for bounding box regression. Keypoint regressions are handled using smooth L1 loss, with ReLU applied to \((u,v)\) and no activation for the other parameters. Despite evaluating the use of sigmoid function for \(\arg\min_u\) and \(\arg\min_v\), we found no significant differences and thus maintain these regressions as real-valued, applying a threshold of 0.5 for binary assignment. Given that BEV representation is invariant to perspective, we adopt a single size range, unlike the five ranges used in \cite{tian2019fcos}. For post-processing, we apply standard non-maximum suppression (NMS) to horizontal bounding boxes, in contrast to the rotated NMS employed by CenterPoint \cite{yin2021center}.

While focal loss effectively addresses the class imbalance issue in single-stage object detectors, we introduce two key modifications to further enhance performance. First, we normalize the classification loss within each class and then aggregate these per-class losses to compute the total classification loss. Additionally, we implement an objectness head to perform binary classification of foreground and background BEV pixels. The foreground indices generated by the objectness head are used to filter bounding box, regression, and centerness losses. This modification allows the model to learn these values even if the classification label is incorrect, provided the pixel is classified as an object. Our experiments demonstrate that these two modifications significantly improve detection performance and reduce the necessity for techniques such as class-balanced grouping and sampling (CBGS) \cite{zhu2019class}.

Although we have selected FCOS as our object detector, the proposed representation is versatile and can accommodate many off-the-shelf 2D object detectors for BEV-based object detection. We support this claim by also implementing the RQR3D to a transformer based object detector.

\subsection{Simplified Radar Processing Backbone}
\label{radar_bb}

While a typical LiDAR sensor can generate highly detailed point clouds consisting of millions of points per second \cite{hunt2024radcloud}, an average automotive radar sensor typically generates up to a few thousand points per scan \cite{schumann2021radarscenes}. Beyond 5-10 meters, one can typically observe only a few points per object. Although radar points are denser on closer objects, the abundance of image-based features reduces the need for this additional information. Motivated by these observations, we hypothesize that voxel grouping of the radar PC becomes unnecessary. Instead, we first project radar points into a higher-dimensional space using two feedforward layers and then map them onto the \(2N \times 2N\) BEV grid using their inherent depth and lateral coordinates. We employ a simple overwrite strategy to retain the last point mapped to each BEV voxel. The radar features are subsequently processed by a lightweight 2D dense convolutional backbone, downsampled to \(N \times N\), and concatenated with the projected image features. Although sparse convolutions are widely used in recent camera-radar 3D perception literature, 2D dense convolutions are shown to perform better than voxelized or pillarized radar PC \cite{scheiner2021object}. Moreover, the inherent 2D organization of radar measurements provides a natural parameterization that supports geometrically consistent processing with standard 2D dense convolutional networks. Our findings indicate that employing a dense 2D convolutional architecture offers improved runtime performance compared to SotA methods and enhanced edge device deployment compatibility, without sacrificing detection accuracy.

\subsection{Overall Model Architecture}

The multi-view camera images are initially processed by an image encoder. In our experiments, we employ three alternative backbone architectures: ResNet50 (R50), ResNet101 (R101) \cite{ResNet}, RegNetY-8GF (R-8GF) \cite{RegNet} with BiFPN \cite{tan2020efficientdet}, and InternImage \cite{Intern}. We pretrained the ResNet and RegNet backbones using nuImages subset of nuScenes \cite{NuScenes}. We project the feature levels at /8, /16, and /32 resolutions into the BEV representation using Lift-Splat \cite{LSS}, incorporating the depth distribution module from BEVDepth \cite{li2023bevdepth}. Unlike BEVDepth, we do not utilize any supervision provided by the LiDAR PC. Additionally, we conducted ablation studies with the simpler depth distribution module of Lift-Splat \cite{LSS}, as well as IPM and MLP-based projections \cite{MLP}, whose results are provided in the Appendices. The radar PC is processed using our proposed simplified radar processing backbone, and the resulting radar features are concatenated with the image features. Features from previous frames are warped to the current frame using egomotion and then concatenated along the channel dimension, similar to BevDet4D \cite{huang2022bevdet4d}. We do not use the gradients produced by the previous frames to update the image encoder. The resulting spatio-temporal BEV features are processed by a ResNet-like backbone, as proposed in \cite{huang2021bevdet}, with additional residual connection and dropout. These features are then shared among task-specific heads. We utilize the proposed 3D object detection head without any class-specific adjustments based on dataset heuristics; in other words, all classes use the same regression heads. However, during training, we double the sizes of ground-truth boxes for pedestrian and traffic cone classes. During evaluation, we halve the box sizes predicted by the model for these two classes. Additionally, we employ an auxiliary segmentation head similar to the one used in Fiery \cite{hu2021fiery}, which learns a semantic segmentation mask, offset mask, and centerness mask, and derives instance masks using these three. We implement BEV augmentations such as flipping, rotating, and zooming-in. We also utilize an auxiliary 2D object detection head based on FCOS \cite{tian2019fcos},  which takes the image features as input.

\section{Experiments}
\label{secexp}

\subsection{Experimental setup}

We conducted our experiments on nuScenes dataset \cite{NuScenes}, a widely adopted benchmark for 3D object detection in autonomous driving research. For model evaluation, we used the official nuScenes metrics: mean Average Precision (mAP), nuScenes Detection Score (NDS), mean Average Translation Error (mATE), mean Average Scale Error, and mean Average Orientation Error (mAOE), with their default settings. All training details, including learning rate schedules, optimization settings and data augmentation techniques are provided in the Appendices for reproducibility.

\subsection{Comparison to the state of the art}

Table \ref{tab:detector_perf_filtered} provides a thorough comparison of RQR3D against state-of-the-art (SotA) methods on the nuScenes validation set, considering two settings of different input resolutions and backbone configurations. RQR3D consistently outperforms previous camera-only and camera-radar fusion approaches in NDS and mATE metrics across these setting. All comparisons are conducted under fair conditions, without the use of CBGS, test-time augmentation (TTA), or temporal cues from future frames, thereby highlighting the efficacy of our architecture under standard evaluation conditions. Table \ref{tab:nuscenes_performance} further evaluates model performance on the nuScenes test set. RQR3D achieves NDS (67.5), mAP (59.7) and mATE (0.356). This demonstrates the robustness and generalizability of our approach across both validation and test benchmarks. 

Furthermore, we evaluate the RQR3D head in a camera-only setting. Since our baseline is based on  BEVDepth, the major difference in our architecture is the replacement of CenterPoint-based detection head with RQR3D. This modification yields an 8.2\% improvement in NDS and a 15.1\% increase in mAP over the BEVDepth baseline as shown in Table \ref{tab:detector_perf_filtered}. Moreover, to demonstrate a minimal scenario, in Table \ref{tab:singleframe_comparison}, we compare single frame performance of RQR3D with other methods whose relevant results are publicly available.

\begin{table*}[ht]
\caption{Comparison of 3D object detection performance on the nuScenes validation set. Unless indicated, obtained without CBGS, TTA and future frames. \textdagger~indicates training with CBGS *~with Future Frames}
\renewcommand{\arraystretch}{0.98} 
\centering
\resizebox{0.81\textwidth}{!}{%
\begin{tabular}{l|ccc|c@{\hskip 6pt}c@{\hskip 6pt}c@{\hskip 6pt}c@{\hskip 6pt}c@{\hskip 6pt}c@{\hskip 6pt}c}
\toprule
\textbf{Method} & \textbf{Input} & \textbf{Backbone} & \textbf{Image Size} & \textbf{NDS} & \textbf{mAP} & \textbf{mATE} & \textbf{mASE} & \textbf{mAOE} & \textbf{mAVE} & \textbf{mAAE} \\
\midrule
PETRv2 \cite{liu2023petrv2} & C & R50 & 704 x 256 & 45.6 & 34.9 & 0.700 & 0.275 & 0.580 & 0.437 & \textbf{0.187} \\
BEVDepth \cite{li2023bevdepth} & C & R50 \textdagger & 704 x 256 & 47.5 & 35.1 & 0.629 & \textbf{0.267} & 0.479 & 0.428 & 0.198 \\
BEVFormerv2 \cite{li2024bevformer} & C & R50 & - & 49.8 & 38.8 & 0.679 & 0.276 & \textbf{0.417} & 0.403 & \textbf{0.189} \\ 
\textbf{RQR3D (Ours)} & C & R50 & 704 x 256 & 51.4 & 40.4 & 0.586 & 0.298 & 0.485 & 0.297 & 0.212 \\
BEVFormerv2 \cite{li2024bevformer} & C & R50 * & - & 52.9 & 42.3 & 0.618 & 0.273 & \textbf{0.413} & 0.333 & \textbf{0.188} \\ 
CRN \cite{CRN}         & C+R & R50  & 704 x 256 & 56.0 & 49.0 & 0.487 & 0.277 & 0.542 & 0.344 & 0.197 \\
RayDN \cite{RayDN} & C   & R50  & 704 x 256 & 56.3 & 46.9 & 0.579 & \textbf{0.264} & 0.433 & 0.256 & \textbf{0.187} \\
CRT-Fusion \cite{CRT} & C+R & R50  & 704 x 256 & 57.2 & 50.0 & 0.499 & 0.277 & 0.531 & 0.261 & 0.192 \\
HyDRa  \cite{Hydra}     & C+R & R50  & 704 x 256 & 58.5 & 49.4 & 0.463 & \textbf{0.268} & 0.478 & 0.227 & \textbf{0.182} \\
RaCFormer (12fps) \cite{racformer} & C+R & R50  & 704 x 256 & 58.8 & 51.0 & - & - & - & - & - \\
RaCFormer \cite{racformer} & C+R & R50  & 704 x 256 & \textbf{61.3} & \textbf{54.1} & 0.478 & \textbf{0.261} & 0.428 & \textbf{0.213} & \textbf{0.180} \\
\textbf{RQR3D (Ours)}        & C+R & R50  & 704 x 256 & 59.2 & 50.7 & \textbf{0.437} & 0.289 & 0.435 & 0.232 & 0.193 \\

\midrule
BEVDepth \cite{li2023bevdepth} & C & R101 \textdagger & 1408 x 512& 53.5 & 41.2 & 0.565 & \textbf{0.266} & 0.358 & 0.331 & 0.190 \\
PETRv2 \cite{liu2023petrv2} & C & R101 & 1600 x 640 & 52.4 & 42.1 & 0.681 & \textbf{0.267} & 0.357 & 0.377 & 0.186 \\
SOLOFusion \cite{SOLO} & C   & R101 \textdagger & 1408 x 512 & 54.4 & 47.2 & 0.518 & 0.275 & 0.604 & 0.310 & 0.210 \\
CRN  \cite{CRN}       & C+R & R101 & 1408 x 512 & 59.2 & 52.5 & 0.460 & 0.273 & 0.443 & 0.352 & \textbf{0.180} \\
RayDN  \cite{RayDN}     & C   & R101 & 1408 x 512 & 60.4 & 51.8 & 0.541 & \textbf{0.260} & \textbf{0.315} & 0.236 & 0.200 \\
HyDRa    \cite{Hydra}   & C+R & R101 & 1408 x 512 & 61.7 & 53.6 & \textbf{0.416} & \textbf{0.264 }& 0.407 & 0.231 & \textbf{0.186} \\
CRT-Fusion \cite{CRT} & C+R & R101 & 1408 x 512 & \textbf{62.1} & 55.4 & 0.425 & \textbf{0.264} & 0.433 & 0.237 & 0.193 \\
RICCARDO \cite{riccardo} & C+R & R101 * & 1408 x 512 & \textbf{62.2} & 54.4 & 0.481 & \textbf{0.266} & 0.325 & 0.237 & 0.189 \\
RaCFormer \cite{racformer} & C+R & R101 & 1408 x 512 & \textbf{63.0} & \textbf{57.3} & 0.476 & \textbf{0.261} & 0.428 & \textbf{0.213} & \textbf{0.180} \\
\textbf{RQR3D (Ours)}        & C+R & R101 & 1408 x 512 & \textbf{62.2} & 54.7 & \textbf{0.417 }& 0.281 & 0.381 & 0.228 & 0.188 \\

\bottomrule
\end{tabular}
\label{tab:detector_perf_filtered}
}
\end{table*}

\begin{table*}[ht]
\caption{Comparison of 3D object detection performance on the nuScenes test set. Unless indicated, obtained without CBGS and TTA. \textdagger~indicates training with CBGS \textdaggerdbl~with TTA *~with Future Frames}
\centering
\resizebox{0.81\textwidth}{!}{%
\begin{tabular}{l|cc|c@{\hskip 6pt}c@{\hskip 6pt}c@{\hskip 6pt}c@{\hskip 6pt}c@{\hskip 6pt}c@{\hskip 6pt}c}
\toprule
\textbf{Method} & \textbf{Input} & \textbf{Backbone} & \textbf{NDS} & \textbf{mAP} & \textbf{mATE} & \textbf{mASE} & \textbf{mAOE} & \textbf{mAVE} & \textbf{mAAE} \\
\midrule
DETR3D \cite{wang2022detr3d} & C & V2-99 & 47.9 & 41.2 & 0.641 & 0.255 & 0.394 & 0.845 & 0.133 \\
PETRv2 \cite{liu2023petrv2} & C & V2-99 & 58.2 & 49.0 & 0.561 & 0.243 & 0.361 & 0.343 & 0.120 \\
BEVDepth \cite{li2023bevdepth}     & C   & ConvNeXt-B \textdagger      & 60.9 & 52.0 & 0.445 & 0.243 & 0.352 & 0.347 & 0.127 \\
SOLOFusion \cite{SOLO}  & C   & ConvNeXt-B \textdagger      & 61.9 & 54.0 & 0.453 & 0.257 & 0.376 & 0.276 & 0.148 \\
CRN \cite{CRN} & C+R & ConvNeXt-B \textdaggerdbl & 62.4 & 57.5 & 0.416 & 0.264 & 0.456 & 0.365 & 0.130 \\
SparseBEV \cite{SparseBEV}   & C   & V2-99           & 63.6 & 55.6 & 0.485 & 0.244 & 0.332 & 0.246 & 0.117 \\
StreamPETR \cite{StreamPETR}  & C   & V2-99           & 63.6 & 55.0 & 0.493 & 0.241 & 0.343 & 0.243 & 0.123 \\
RCBEVDet \cite{RCB}    & C+R & V2-99           & 63.9 & 55.0 & 0.390 & \textbf{0.234} & 0.362 & 0.259 & 0.113 \\
HyDRa  \cite{Hydra}      & C+R & V2-99           & 64.2 & 57.4 & 0.398 & 0.252 & 0.424 & 0.250 & 0.122 \\
RayDN   \cite{RayDN}     & C   & V2-99           & 64.5 & 56.5 & 0.461 & 0.241 & 0.322 & 0.239 & \textbf{0.114} \\
CRT-Fusion \cite{CRT}   & C+R & ConvNeXt-B      & 64.9 & 58.3 & 0.365 & 0.261 & 0.405 & 0.262 & 0.132 \\
RaCFormer \cite{racformer} & C+R & V2-99 & 65.9 & 59.2 & 0.407 & 0.244 & 0.345 & 0.238 & 0.132 \\
RICCARDO \cite{riccardo} & C+R & V2-99 * & \textbf{69.5} & \textbf{63.0} & 0.363 & 0.240 & 0.311 & \textbf{0.167} & 0.118 \\
RQR3D (Ours)      & C+R & ConvNeXt-B   & 66.6 & 59.2 & \textbf{0.361} & 0.252 & 0.341 & 0.218 & 0.134 \\
RQR3D (Ours)      & C+R & InternImage-B   & 67.5 & 59.7 & \textbf{0.356} & 0.245 & \textbf{0.298} & 0.204 & 0.127 \\
\bottomrule
\end{tabular}
\label{tab:nuscenes_performance}
}
\end{table*}

\begin{table}[ht]
\caption{Comparison of single frame performance for 704x256 resolution, R50 backbone \cite{ResNet} over CRN \cite{CRN} and CRT-Fusion \cite{CRT}.}
\centering
\begin{tabular}{l|c@{\hskip 6pt}c@{\hskip 6pt}c@{\hskip 6pt}c@{\hskip 6pt}c}
\toprule
\textbf{Method} & \textbf{NDS} & \textbf{mAP} & 
\textbf{mATE} & \textbf{mAOE} \\
\midrule
CRN          & 50.3 & 42.9 & 0.519 & 0.577 \\
CRT-Fusion   & 50.4 & 44.4 & 0.515 & 0.598 \\
\midrule
RQR3D (Ours) & 53.9 & 44.0 & 0.497 & 0.453 \\
\bottomrule
\end{tabular}
\label{tab:singleframe_comparison}
\end{table}

\textbf{General Applicability of RQR3D.} To validate the general applicability of RQR3D across different detection frameworks, we implement RQR3D on a DETR3D-sytle camera-radar fusion architecture (single frame, 704×256 input, R50 backbone) and compare it with FUTR3D \cite{futr3d}, which utilizes a similar architecture but adopting DETR as 3D object detection head. As reported in Table \ref{tab:futr3d_comparison}, RQR3D provides consistent improvements across different detection framework, supporting our claim about its versatility.

\begin{table}[t]
\centering
\caption{Comparison of transformer version of RQR3D to a transformer-based baseline method FUTR3D for single frame training.}
\begin{tabular}{lcccc}
\toprule
\textbf{Method} & \textbf{Modality} & \textbf{NDS} & \textbf{mAP} \\
\midrule
FUTR3D \cite{futr3d} & C+R & 51.1 & 39.9 \\
RQR3D (transformer)   & C+R & 51.1 & 42.0 \\
\bottomrule
\end{tabular}

\label{tab:futr3d_comparison}
\end{table}

\textbf{Inference Time Analysis.} 
The RQR3D model (single frame, 704×256 input, R50 backbone, FP16) consumes 2.6 GB of memory and runs at 14.8 fps on an NVIDIA A5000 GPU. For a fair comparison with the models reporting their inference time on NVIDIA RTX 3090, we scale our inference time by a factor of 1.28 (derived from the F16 throughput of both GPUs) to estimate the equivalent fps on RTX 3090 as 19 for the single frame version, and as 17.4 for the temporal version whose nuScenes validation set performance reported in Table \ref{tab:detector_perf_filtered}. Table \ref{tab:runtime_comparison} compares RQR3D with other approaches through their estimated RTX 3090 equivalent performances.


\begin{table}[t]
\centering
\small
\caption{Runtime comparison across camera-radar fusion models. Number of temporal frames used are given in parentheses. Values for models with *~are calculated as equivalent.}
\begin{tabular}{l@{\hskip 3pt}c@{\hskip 5pt}c@{\hskip 3pt}c@{\hskip 3pt}c@{\hskip 3pt}c@{\hskip 3pt}c}
\toprule
\textbf{Method} & \textbf{Radar Bb.} & \textbf{Head} & \textbf{Total} & \textbf{FPS} & \textbf{GPU RAM} \\
& (ms)                & (ms)          & (ms)           &   & (GB)             \\
\midrule
CRN (0t)                & 7.6 & 7.9 & 55.6 & 18.0 & 4.2 \\
CRN (7t)                & --     & --     & 273  & 3.7  & 4.3 \\
CRT-Fusion (0t)         & --     & --     & 57.1 & 17.5 & 3.6 \\
CRT-Fusion (6t)         & 13.9 & 8.9 & 67.1 & 14.9 & 3.7 \\
RICCARDO* (7t)          & 48.0 & --     & 108  & 9.3  & -- \\
RaCFormer (4t)    & --     & --     & --      & 12.0 & -- \\
RaCFormer* (7t)         & --     & --     & --      & 1.5  & -- \\
RQR3D* (0t)              & 5.7 & 7.0 & 52.6 & 19.0 & 2.6 \\
RQR3D* (7t)              & 5.7 & 7.0 & 57.2 & 17.4 & 2.7 \\
\bottomrule
\end{tabular}
\label{tab:runtime_comparison}
\end{table}

\subsection{Ablations}
We have conducted the following procedures at an image resolution of 1408×512 using the RegNet-Y 8GF backbone \cite{RegNet} with a Bidirectional Feature Pyramid Network (BiFPN Neck) \cite{tan2020efficientdet}, aligned with the final network target configuration.


\textbf{RQR3D vs. CenterPoint.} In Table \ref{tab:rbb_temp_abl}, we compare RQR3D with CenterPoint \cite{yin2021center} under various settings, while keeping the rest of the network identical for both detection heads. In the first setting, no previous frames are used, and the radar PC is directly mapped to BEV without employing the proposed radar backbone. Subsequently, we enable the proposed radar backbone and temporality, respectively. The proposed radar backbone provides an improvement of 1.4 mAP and 2.3\% mATE. The inclusion of temporal frames further enhances the mAP by 5.4 points and mATE by an additional 10.6\%. RQR3D consistently outperforms CenterPoint head in all scores across all three settings, indicating the superiority of the proposed quadrilateral representation over the angle-based representation used in CenterPoint head.

\textbf{Objectness Head, Augmentations and Auxiliary Segmentation Task.} Our baseline model employs the objectness head, as introduced in Section \ref{3dhead}, to address class imbalance issues. It also incorporates image and BEV augmentations to ensure generalization and robustness, and utilizes an auxiliary segmentation task to support the learning of the main 3D object detection task. In Table \ref{tab:seg_aug_abl}, we systematically remove these components one by one to evaluate their individual contributions to the overall model performance. We observed that augmentations enhance the precision score by 3.0 mAP, while the segmentation head substantially improves the mATE score by 5.3\%. The most notable overall improvement is attributed to the objectness head, which increases the mAP by 3.9 and the mATE by 6.5\%. This observation empirically demonstrates that utilizing the positive-negative indices generated by the objectness head, significantly enhances the performance of a one-stage structure, eliminating the need for a two-stage architecture.

\begin{table}[ht]
\centering
\small
\renewcommand{\arraystretch}{0.99} 
\caption{Ablation of RQR3D and CenterPoint detection heads with options of radar backbone and temporal frame utilization.}
\begin{tabular}{c@{\hskip 4pt}c|c@{\hskip 4pt}|c@{\hskip 4pt}c@{\hskip 4pt}c}
\toprule
\textbf{Radar Bb.} & \textbf{Tem. Fr.} & \textbf{Det. Head} & \textbf{mAP} & \textbf{mATE} & \textbf{mAOE} \\
\midrule
\multirow{2}{*}{\ding{55}} & \multirow{2}{*}{\ding{55}} 
  & CenterPoint & 47.2 & 0.507 & 0.561 \\
 &  & RQR3D & 48.5 & 0.489 & 0.375 \\
\midrule
\multirow{2}{*}{\ding{51}} & \multirow{2}{*}{\ding{55}} 
  & CenterPoint & 48.0 & 0.496 & 0.622 \\
 &  & RQR3D & 49.9 & 0.478 & 0.344 \\
\midrule
\multirow{2}{*}{\ding{55}} & \multirow{2}{*}{\ding{51}} 
  & CenterPoint & 53.3 & 0.451 & 0.578 \\
 &  & RQR3D & 53.8 & 0.445 & 0.400 \\
\midrule
\multirow{2}{*}{\ding{51}} & \multirow{2}{*}{\ding{51}} 
  & CenterPoint & 53.4 & 0.448 & 0.562 \\
 &  & RQR3D & 55.3 & 0.432 & 0.389 \\
\bottomrule
\end{tabular}
\label{tab:rbb_temp_abl}
\end{table}
\begin{table}[ht]
\centering
\small
\renewcommand{\arraystretch}{0.9} 
\setlength{\tabcolsep}{6pt}
\caption{Ablation on auxiliary objectness and segmentation heads and augmentations for RQR3D.}
\begin{tabular}{c@{\hskip 6pt}c@{\hskip 6pt}c|c@{\hskip 6pt}c}
\toprule
\textbf{Obj. Head} & \textbf{Augmentations} & \textbf{Segm. Head} & \textbf{mAP} & \textbf{mATE} \\
\midrule
\ding{55} & \ding{51} & \ding{51} & 44.6 & 0.521 \\
\ding{51} & \ding{55} & \ding{51} & 45.5 & 0.485 \\
\ding{51} & \ding{51} & \ding{55} & 47.2 & 0.515 \\
\midrule
\ding{51} & \ding{51} & \ding{51} & 48.5 & 0.489 \\
\bottomrule
\end{tabular}
\label{tab:seg_aug_abl}
\end{table}

\section{Discussions and Conclusion}
In this paper, we present the Restricted Quadrilateral Representation for regression targets in 3D object detection. RQR3D defines the smallest horizontal bounding box around the oriented box and uses two offsets to ensure the representation of only rectangular cuboids with right angles. This approach transforms the problem into a keypoint regression task, which is more amenable to learning with translation-invariant convolutions.  RQR3D focuses on the regression target representation and is compatible with different 3D object detection methods. We adapt an anchor-free single-stage 2D object detection method for BEV-based 3D object detection. Also, we propose a lightweight radar fusion backbone that eliminates the need for voxel grouping and processes the BEV-mapped point cloud using 2D convolutions.

RQR3D uses the shortest offset from two corners, and indicates the chosen offset using a binary-valued target \((\arg\min_u, arg\min_v)\). This strategy mitigates the loss discontinuity issues. However, inaccuracies in estimating \((\arg\min_u, arg\min_v)\) can lead to errors in corner placement or size estimation, resulting in a slight elevation in mASE. While a simple \(w \le l\) check and edge swap strategy can alleviate this issue, future research should focus on enhancing the estimation accuracy of \((\arg\min_u, arg\min_v)\).

While applying standard NMS to horizontal bounding boxes is both simpler and faster, it is susceptible to erroneous suppressions in regions with high object density due to potential intersections of the encapsulating boxes. Although this issue has not been observed in our visualizations it has the effect of reducing mAP score slightly. Implementing a rotated version of NMS could be explored in future work to enhance robustness and ensure safer detection performance.

\bibliographystyle{ieeetr} 
\bibliography{kutup} 



\maketitlesupplementary
\setcounter{page}{1}
\section{Obtaining the 3D bounding box parameters}

Evaluation tool from nuScenes \cite{NuScenes} requires 3D bounding boxes to be defined as \((x_\text{ctr}, y_\text{ctr}, z_\text{ctr}, w, l,h, \theta)\). 
In this section, we provide the details about how we obtain this representation using RQR3D outputs, \((x_{\text{min}}, y_{\text{min}}, x_{\text{max}}, y_{\text{max}})\) and \((u, v, \arg\min_u, \arg\min_v\, d_x, d_y, z_{\text{ctr}}, h)\).

Recalling that \(u\) represents the minimum of the offsets on x-axis, \(v\) represents the minimum of the offsets on y-axis, and \((\arg\min_u, \arg\min_v)\) represent their indices, respectively. We first reassign \((u,v)\) with respect to \((x_{\min}, y_{\min})\) as follows: 

\begin{equation}
u = 
\begin{cases}
u, & \text{if } \arg\min_u < 0.5 \\
(x_{\text{max}} - x_{\text{min}}) - u, & \text{otherwise}
\end{cases}
\end{equation}

\begin{equation}
v = 
\begin{cases}
v, & \text{if } \arg\min_v < 0.5 \\
(y_{\text{max}} - y_{\text{min}}) - v, & \text{otherwise}
\end{cases}
\end{equation}

Then, we calculate \((w,l)\) as
\begin{equation}
A=\sqrt{(x_{\max}-u)^2+v^2},\quad
B=\sqrt{u^2+(y_{\max}-v)^2}.
\end{equation}
\begin{equation}
w=\begin{cases}A,& d_x d_y \ge 0\\ B,& \text{otherwise}\end{cases},\qquad
l=\begin{cases}B,& d_x d_y \ge 0\\ A,& \text{otherwise}\end{cases}.
\end{equation}



and \((x_\text{ctr}, y_\text{ctr})\) as 

\begin{equation}
    x_\text{ctr} = \frac{x_\text{min} + x_\text{max}}{2}, \quad
    y_\text{ctr} = \frac{y_\text{min} + y_\text{max}}{2}.
\end{equation}

\(z_\text{ctr}, h\) can be used directly as regressed. Finally, we obtain \(\theta\) using \((d_x,d_y\)) as follows.

\begin{equation}
\theta = \arctantwo(d_y, d_x).
\end{equation}

Similarly, one can obtain the orientation using \((u,v,x_\text{max} - u, y_\text{max}-v)\) in a similar fashion to \((w,l)\) calculation. However, in this case the angular range would be \(180^o\). Although the sign of \((d_x,d_y)\) can be used to find the correct quadrant and extend the angular range to \(360^o\), we observe that calculating the orientation directly from \((d_x,d_y)\) yields lower angular error. 

In nuScenes convention, the corners of \(\mathcal{B}\) are in the order of \textit{front-left, front-right, rear-right, rear-left}. Let \(V_1,V_2,V_3,V_4\) be the the bottom corners of \(\mathcal{B}\) in this given order.

\begin{equation}
\label{v1}
    V_1 = 
    \begin{cases}
        (x_{\text{max}},\ y_{\text{min}} + v,\ z_\text{bottom}), & \text{if } d_x \geq 0 \land d_y \geq 0 \\
        (x_{\text{min}} + u,\ y_{\text{min}},\ z_\text{bottom}), & \text{if } d_x \geq 0 \land d_y < 0 \\
        (x_{\text{max}} - u,\ y_{\text{max}},\ z_\text{bottom}), & \text{if } d_x < 0    \land d_y \geq 0 \\
        (x_{\text{min}},\ y_{\text{max}} - v,\ z_\text{bottom}), & \text{if } d_x < 0    \land d_y < 0 
    \end{cases}
\end{equation}

\begin{equation}
\label{v2}
    V_2 = 
    \begin{cases}
        (x_{\text{max}} - u,\ y_{\text{max}},\ z_\text{bottom}), & \text{if } d_x \geq 0 \land d_y \geq 0 \\
        (x_{\text{max}},\ y_{\text{min}} + v,\ z_\text{bottom}), & \text{if } d_x \geq 0 \land d_y < 0 \\
        (x_{\text{min}},\ y_{\text{max}} - v,\ z_\text{bottom}), & \text{if } d_x < 0    \land d_y \geq 0 \\
        (x_{\text{min}} + u,\ y_{\text{min}},\ z_\text{bottom}), & \text{if } d_x < 0    \land d_y < 0 
    \end{cases}
\end{equation}

\begin{equation}
\label{v3}
    V_3 = 
    \begin{cases}
        (x_{\text{min}},\ y_{\text{max}} - v,\ z_\text{bottom}), & \text{if } d_x \geq 0 \land d_y \geq 0 \\
        (x_{\text{max}} - u,\ y_{\text{max}},\ z_\text{bottom}), & \text{if } d_x \geq 0 \land d_y < 0 \\
        (x_{\text{min}} + u,\ y_{\text{min}},\ z_\text{bottom}), & \text{if } d_x < 0    \land d_y \geq 0 \\
        (x_{\text{max}},\ y_{\text{min}} + v,\ z_\text{bottom}), & \text{if } d_x < 0    \land d_y < 0 
    \end{cases}
\end{equation}

\begin{equation}
\label{v4}
    V_4 = 
    \begin{cases}
        (x_{\text{min}} + u,\ y_{\text{min}},\ z_\text{bottom}), & \text{if } d_x \geq 0 \land d_y \geq 0 \\
        (x_{\text{min}},\ y_{\text{max}} - v,\ z_\text{bottom}), & \text{if } d_x \geq 0 \land d_y < 0 \\
        (x_{\text{max}},\ y_{\text{min}} + v,\ z_\text{bottom}), & \text{if } d_x < 0    \land d_y \geq 0 \\
        (x_{\text{max}} - u,\ y_{\text{max}},\ z_\text{bottom}), & \text{if } d_x < 0    \land d_y < 0 
    \end{cases}
\end{equation}

where \(z_{bottom}=z_\text{ctr}-\nicefrac{h}{2}\). Then, let \(V_5,V_6,V_7,V_8\) be the the top corners of \(\mathcal{B}\) in the same order. Similarly, their coordinates can be obtained using Eq. \ref{v1}-\ref{v4} replacing \(z_\text{bottom}\) with \(z_{top}=z_\text{ctr}+\nicefrac{h}{2}\).

\section{Implementation Details}
All models are trained using a batch size of 8 and an initial learning rate of \(7.5 \times 10^{-5}\), optimized using the Adam optimizer. Training is conducted over 20 epochs with a multi-step learning rate schedule: the learning rate is reduced by a factor of 10 at epochs 15 and 18. The bird’s eye view (BEV) representation covers a spatial extent of 100 meters by 100 meters, corresponding to a 50-meter range in all directions from the ego vehicle. The depth prediction network estimates depth within the range of 2 to 50 meters, discretized into 48 bins. All experiments are carried out on four NVIDIA A5000 GPUs.

We apply standard BEV augmentations as commonly used in the literature \cite{huang2022bevdet4d}. Horizontal and vertical flipping of the BEV, along with corresponding image adjustments, are applied independently with a probability of 0.5. As a result, each of the four flipping configurations (none, horizontal only, vertical only, and both) occurs with equal probability of 0.25. We do not apply random scaling or image-space rotations. Instead, to augment the BEV feature representation, we apply discrete BEV rotations. With a probability of 0.5, the BEV is rotated by an angle randomly selected from the set [45, 90, 135, 180, 225, 270, 315] otherwise, no rotation is applied. These rotations affect only the projection of sensor information (camera and radar) into BEV space and do not modify the raw images or radar PC data. Additionally, we apply BEV scaling with a probability of 0.3. This augmentation reduces the spatial coverage from 50 meters to 25 meters, thereby increasing the resolution of nearby regions and enabling more precise detection of objects in close proximity.

\section{Additional Experiments}
\subsection{Projection Methods} 
In Table \ref{tab:projection_abl}, we evaluate the contribution of various projection methods to the overall performance. Our baseline model employs Lift-Splat projection with BEVDepth's depth distribution module, denoted as DN. We compare this baseline with three different versions: i) Lift-Splat projection with a simpler depth distribution, denoted as LSS, ii) IPM-based projection, and iii) MLP-based projection. The IPM, LSS, and MLP versions achieve comparable performance with each other, while the DN version yields the highest performance by leveraging the enhanced depth distribution module.

\begin{table}[ht]
\caption{Projection methods.}
\centering
\begin{tabular}{l|ccc}
\toprule
\textbf{Projection} & \textbf{mAP} & \textbf{mATE} & \textbf{mASE} \\
\midrule
LSS  & 54.5 & 0.444 & 0.278 \\
IPM & 54.5 & 0.439 & 0.279 \\
MLP & 54.6 & 0.439 & 0.285\\
\hline
DN (Baseline)  & 55.3 & 0.432 & 0.277\\
\bottomrule
\end{tabular}
\label{tab:projection_abl}
\end{table}

\subsection{Qualitative Comparison between CenterPoint and RQR3D}

\begin{figure}
	\begin{center}
		\centerline{\includegraphics[width=1.0\linewidth]
        {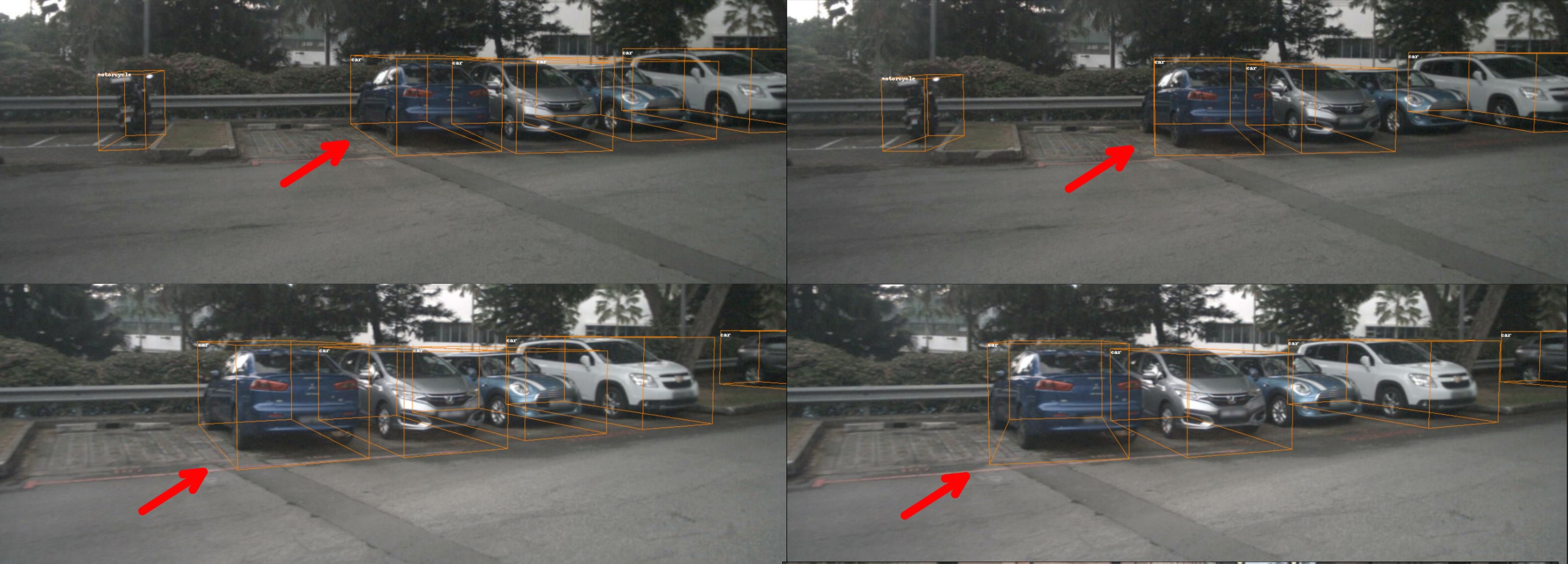}}
		\caption{
        Orientation stability of RQR3D head (left) compared to CenterPoint head \cite{yin2021center} (right)  
        }
		\label{fig:RQRCP}
	\end{center}
	\vskip -0.3 in
\end{figure}

RQR3D head have a superior orientation regression compared to CenterPoint head with angle-based targets. This has been established quantitatively in the main paper. Figure \ref{fig:RQRCP} shows an example where CenterPoint head estimates the orientation drastically different between two consecutive frames, whereas RQR3D head provides a more stable estimation.

\subsection{Inference Time Analysis} 

The RQR3D model (with single frame, 704x256 image resolution and R50 backbone), when executed with FP16 precision, utilizes approximately 2.6 GB of memory—equivalent to 10\% of the available capacity—on the NVIDIA A5000 GPU. Under these conditions, the model achieves an average inference speed of 14.8 frames per second (fps). For comparative purposes, performance estimates are referenced against the NVIDIA RTX 3090 GPU. The A5000 and RTX 3090 offer peak FP16 computational throughputs of 222.2 and 285.5 TFLOPs, respectively. Given that the inference process utilizes only 1.5\% of the system memory and 30\% CPU resources, I/O bottlenecks are negligible, and compute performance becomes the primary scaling factor. Accordingly, inference speed is expected to scale by a factor of approximately 1.28 when transitioning from the A5000 to the RTX 3090, yielding an estimated equivalent performance of 19 fps for RQR3D on the latter. During this inference, the auxiliary segmentation head and 2D object detection heads are removed. Similarly the objectness head outputs are not used during inference, therefore objectness head is also removed.

\end{document}